\documentclass[letterpaper, 10 pt, conference]{ieeeconf}  

\IEEEoverridecommandlockouts                              

\usepackage{tabularx}
\usepackage{amsfonts}
\usepackage{graphicx}
\usepackage{amsmath}
\usepackage{arydshln}
\usepackage{relsize}

\usepackage{subcaption}

\usepackage[hyphens]{url}

\usepackage{xcolor}

\newcommand{\method}{Gen-LSNE }
\title{\LARGE \bf
Generative Low-Shot Network Expansion
}

\author{Adi Hayat \hspace{0.2cm} Mark Kliger* \hspace{0.2cm} Shachar Fleishman*  \hspace{0.2cm} Daniel Cohen-Or
\thanks{ *This work was done while Mark Kliger and Shachar Fleishman were with Intel Corporation.} \\
\hspace{0.5cm} Intel \hspace{1.0cm} Amazon \hspace{1.5cm} Amazon \hspace{1.2cm} Tel-Aviv University \\
\texttt{adi.hayat@intel.com,\{mark.kliger,shacharfl,cohenor\}@gmail.com } \\
}

\begin{document}
\maketitle
\thispagestyle{empty}
\pagestyle{empty}

\begin{abstract}

Conventional deep learning classifiers are static in the sense that they are trained on a predefined set of classes and learning to classify a novel class typically requires re-training. In this work, we address the problem of \emph{Low-Shot network-expansion} learning. We introduce a learning framework which enables expanding a pre-trained ({\em base}) deep network to classify novel classes when the number of examples for the novel classes is particularly small. We present a simple yet powerful {\em hard distillation} method where the base network is augmented with additional weights to classify the novel classes, while keeping the weights of the base network unchanged.

We show that since only a small number of weights needs to be trained, the hard distillation excels in low-shot training scenarios. Furthermore, hard distillation avoids detriment to classification performance on the base classes.
Finally, we show that low-shot network expansion can be done with a very small memory footprint by using a compact generative model of the base classes training data with only a negligible degradation relative to learning with the full training set.

\end{abstract}

\section{INTRODUCTION}

\begin{figure}[h]
\begin{center}
\includegraphics[width=0.85\linewidth]{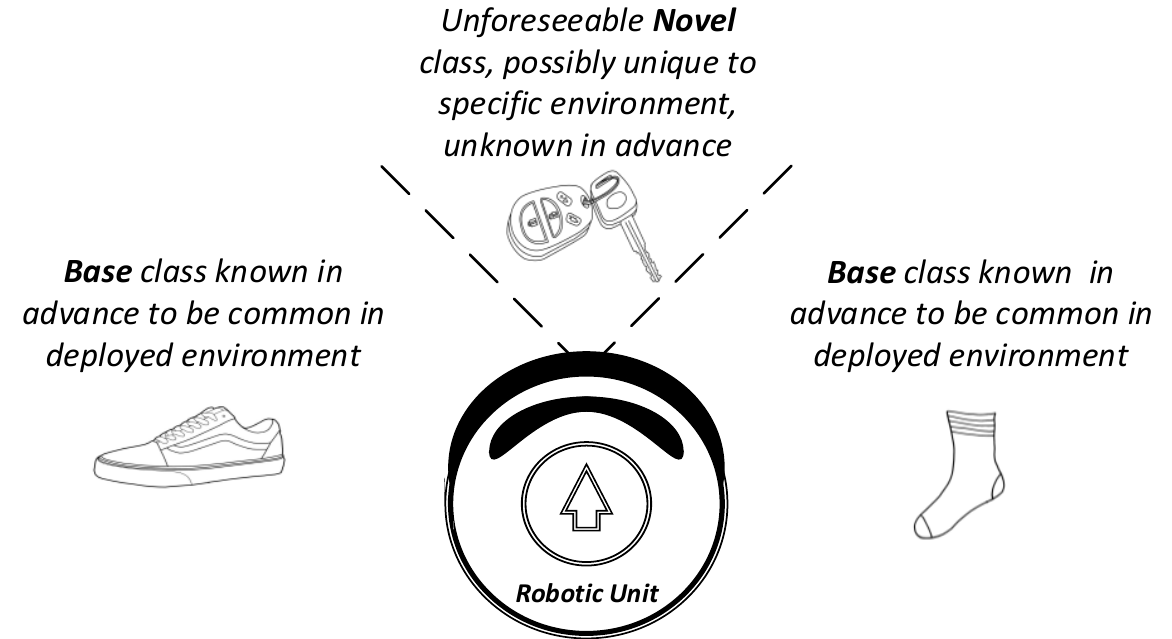}
\end{center}
   \caption{\textbf{Problem Setup}: 
   We desire to be able to adapt the base classifier to the \textbf{Novel} classes in the deployed setting.}
\label{fig:Setup}
\end{figure}

%
In many real-life scenarios, a fast and simple \emph{classifier expansion} is required to extend the set of classes that a deep network can classify. For example, consider a cleaning robot trained to recognize a number of objects. 
After deployment, 
the robot is likely to encounter novel objects which it was not trained to classify.  
It is desired to be able to update and expand the robot classifier to classify novel objects. In such a scenario, the update should be a simple procedure, based on a small collection of images captured in a non-controlled setting. 
This scenario is illustrated in Figure \ref{fig:Setup}: we wish to update and expand a robot classifier to classify novel classes in the deployed setting.

The low-shot network update should be fast and without requiring access to the \emph{entire training set} of previously learned data. A common solution to classifier expansion is \emph{fine-tuning} the network \cite{Kaeding16_FDN}. However fine-tuning requires collecting sufficient examples of the novel classes, in addition to keeping a large amount of {\em base} training data in memory, to avoid 
{\em catastrophic forgetting} \cite{CForget}. In striking contrast, for some tasks, humans are capable of instantly learning novel categories. Using one or only a few training examples, humans are able to learn a novel class, without compromising previously learned abilities or having access to training examples from all previously learned classes.

We consider the classifier expansion problem under the following constraints: 
\begin{enumerate}
 \setlength\itemsep{0em}
\item Low-shot: very few samples of the novel classes are available. 
\item No forgetting: preserving classification performance on the base classes. 
\item Small memory footprint: no access to the base classes training data. 

\end{enumerate}

In this work, we introduce a low-shot network expansion technique, augmenting the capability of an existing (base) network trained on base classes by training additional parameters that enable to classify novel classes. 

To satisfy low-shot along with no-forgetting constraints, we present a {\it hard distillation} framework. 
Distillation in neural networks \cite{distilation-hinton} is a process for training a target network to imitate another network. A loss function is added to the target network so that its output matches the output of the mimicked network. In standard {\it soft distillation}, the trained network is allowed to deviate from the mimicked network. Whereas hard distillation enforces that the output of the trained network for base classes matches the output of the mimicked network as a hard constraint. 
Network expansion with hard distillation yields a larger network, distilling the knowledge of the base network in addition to the augmented capacity to classify novel classes.  
We show that in the case of low-shot (only 1--15 examples of a novel class), hard distillation outperforms soft distillation. 

To maintain a small memory footprint, we refrain from saving the entire training set. Instead, we present a compact generative model, consisting of a collection of generative models fitted in the feature space to each of the base classes. We use a Gaussian Mixture Model (GMM) with a small number of mixtures, and show it inflicts a minimal degradation in classification accuracy. Sampling from the generative GMM model is fast, reducing the low-shot training time and allowing fast expansion of the network. 

We define a benchmark for low-shot network expansion. The benchmark is composed of a series of tests of increasing complexity. 
We perform a comprehensive set of experiments on this challenging benchmark, comparing the performance of the proposed to alternative methods.



\section{Related Works}
\label{sec:realted-work}

A common solution to the class-incremental learning problem is to use a Nearest-Neighbors (NN) based classifier in feature space. 
A significant advantage of an NN-based classifier is that it can be easily extended to classify a novel class, even when only a single example of the class is available ({\it one-shot learning}). However NN-based classifiers keep in the memory significant amount data. \cite{mensink13pami} proposed to use Nearest Class Mean (NCM) classifier, where each class is represented by a single prototype example which is the mean feature vector of all class examples.
One major disadvantage of NCM and NN-based methods is that they are based on a fixed feature representation of the data. To overcome this problem \cite{mensink13pami} proposed to learn a new distance function in the feature space using metric learning.

The Incremental Classifier and Representation Learning (iCaRL) method \cite{icarl} aims to solve the class-incremental learning problem using the Nearest-Mean-of-Exemplars classifier method. Feature representation is updated and the class means are recomputed from a small stored number of representative examples of the base classes. During the feature representation update, the network parameters are updated by minimizing a combined classification and distillation loss. The iCaRL method was introduced as a class-incremental learning method for large training sets. 


In \cite{hariharan2016lowshot} a \emph{Squared Gradient Magnitude} regularization technique was proposed that improves the \emph{fixed} feature representation for low-shot scenarios. They also propose to hallucinate additional training examples from the novel classes. In contrast, we present a method which aims to maximize the performance in low-shot network expansion \emph{given} a fixed representation. 

In {\em Progressive Network} \cite{PNN-Ruso-2017}, new tasks are learned without affecting the performance of old tasks by {\it freezing} the parameters of old tasks and {\it expanding} the network with additional layers to solve new tasks.
{\em Progressive learning} \cite{Venkatesan:2016:NPL} solves the problem of online sequential learning in extreme learning machines (ELM). The purpose of their work is to incrementally learn the last fully-connected layer of the network. 
In \cite{venkatesan2017strategy} was proposed an incremental learning technique which augments the base network with additional parameters in the last fully connected layer to classify novel classes. Similar to iCaRL, it performs soft distillation by learning all parameters of the network. The {\it phantom sampling} for hallucinating data from past distribution modeled with Generative Adversarial Networks was used instead of keeping historical training data.

In this work, we propose a solution that borrows ideas from the freeze-and-expand paradigm, improved feature representation learning, network distillation and modeling past data with a generative model. We propose to expand the last fully connected layer of a base network to classify novel classes. Moreover, the deeper layers may be also expanded to improve the feature representation. However, in contrast to previous methods \cite{icarl,Venkatesan:2016:NPL}, we do not retrain the base network parameters, but only train the expanded parts of the network. 
The extended feature representation is learned from samples of base and novel classes. Finally, in order to avoid keeping all of the historical training data, we use a GMM of the feature space as a generative model for the base classes. 

\section{The proposed method}
\label{sec:the-method}

Assume a deep neural network is trained on $K$ base classes with the full set of training data. This base network can be partitioned into two subnetworks: a feature extraction network and a classification network. The feature extraction network $f_{rep}$ maps an input sample $x$ into a feature representation $v \in\mathbb{R}^N$. The classification network $f_{cls}$ maps feature vectors $v$ into a vector of approximated class posterior probabilities $P\left( {{k}|v} \right)$ which correspond to each one of $K$ classes. The whole network  can be represented as composition of two networks $f_{net} (x) = f_{cls}(f_{rep}(x))$. 

In the following, we discuss how the pre-learned feature representation of feature extraction network can be leveraged to classify additional classes in a low-shot scenario with only relatively minor changes to the classification subnetwork.

\subsection{Expansion of the last layer of classification subnetwork}
\label{single_layer}
First, we discuss how to expand the classification network to classify one additional class. We can expand $f_{cls}$ from a $K$-class classifier into $K+1$ class classifier by adding a new weight vector $w_{K+1}\in\mathbb{R}{^N}$ to the last FC layer. Thus, the $K+1$ class probability is $f_{cls}(v)[K+1] = \frac{1}{Z'}{e^{w_{K+1}^T v}}$, where $Z'$ is a new normalization factor for $K+1$ classes. 
We would like to preserve classification accuracy on the base classes to avoid catastrophic forgetting. To that end, during training we constrain to optimize of the $w_{K+1}$ weights, while the vectors $\left\{ {{w_i}} \right\}_{i = 1}^K$ are kept intact. We refer to this paradigm as {\em hard distillation}. By preserving the base classes weight vectors, we guarantee that as a result of the last classification layer expansion the only new errors that can appear are between the novel class and the base classes, but not among the base classes. Moreover, the small number of newly learned parameters helps avoid over-fitting, which is especially important in low-shot scenarios. 

Similarly, we can expand the classification network to classify more than one novel class.
\begin{figure}
\begin{center}
\begin{tabular}[t]{c}
\includegraphics[width=0.7\linewidth]{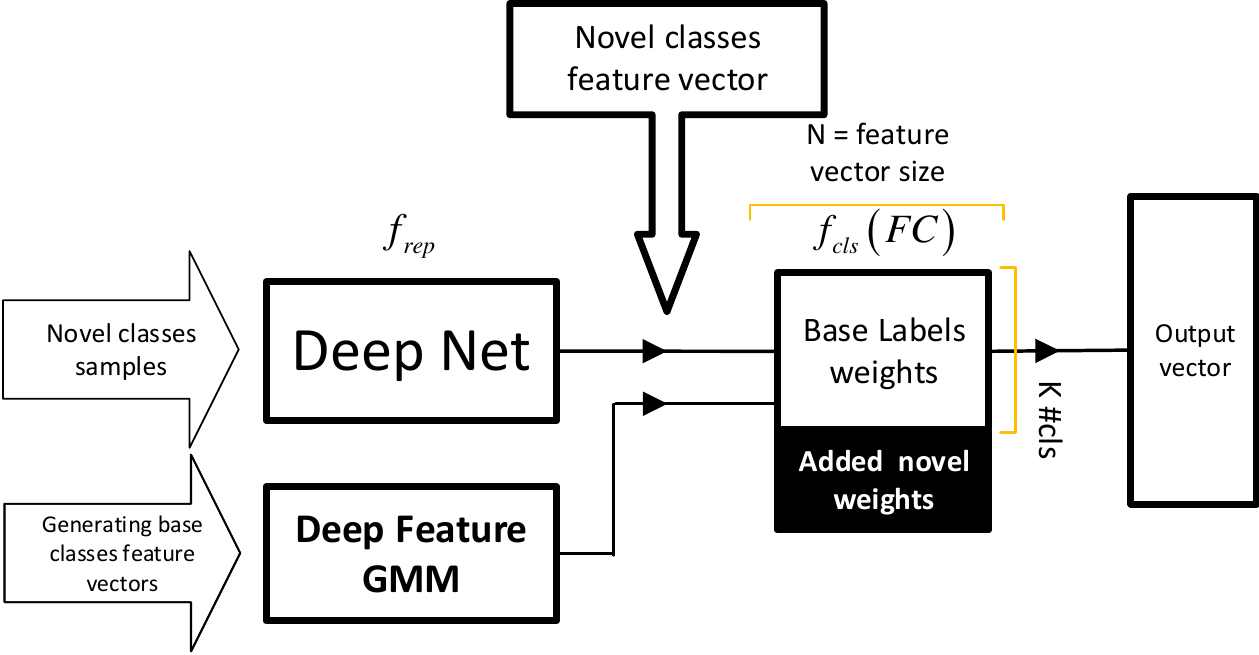} \\
(a) \\
\includegraphics[width=0.7\linewidth]{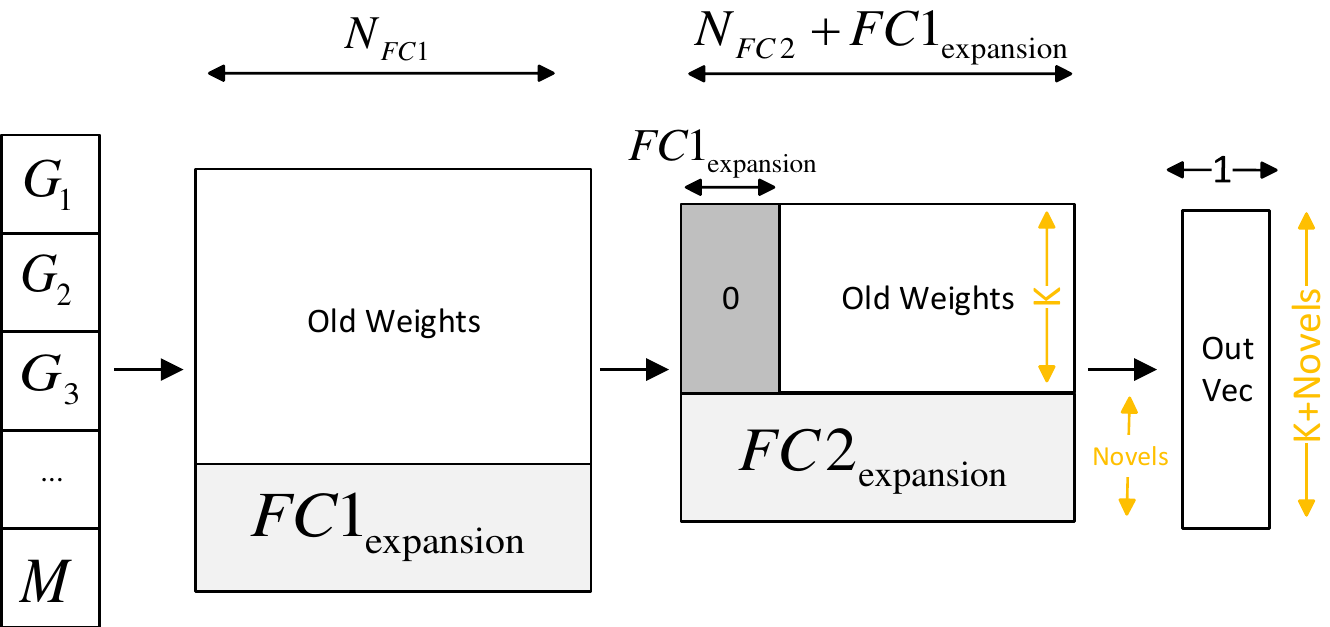} \\
(b)
\end{tabular}
\end{center}
   \caption{(a) The proposed \method overview, generating $f_{rep}$ feature representation of base classes to train the $f_{cls}$ expansion. (b) Training the last two layers, learning shared representation in addition to the per novel class weights expansion: $G_i$ are samples of feature vector generations of base class $i$, $M$ are the novel class feature vector measurements, $N_{FC1}$ are the number of input features to $FC1$, $N_{FC2}$ are the number of input feature to $FC2$ before the expansion.}
\label{fig:Method}
\end{figure}
\subsection{Deep Feature GMM - Generative model for base classes}
\label{Depp_feature_GMM}
Due to the small memory footprint constraint, we are unable to keep the entire training data of the base classes. As an alternative, we can use a generative model of the base classes and during training draw samples from the model. There are  various approaches to this task, such as GAN \cite{goodfellow_generative_2014}, VAE \cite{VAE}, Pixel CNN \cite{PixelCNN}, or conventional methods of non-parametric kernel density estimation \cite{hastie01statisticallearning}. However, it is usually hard to generate \emph{accurate} samples from past learned distributions in the \emph{image} domain, and these methods still require a significant amount of memory to store the model network parameters. Furthermore, since training typically requires thousands of samples, we prefer a generative model that allows fast sampling to reduce the low-shot phase training time. 

In our work, we use the Gaussian Mixture Model (GMM) density estimator as an approximate generative model of the data from the base classes. However, instead of approximating the generative distribution of the \emph{image} data, we approximate a class conditional distribution of its feature representation. Thus, we model a GMM $P\left( {v|{c=k}} \right)=\sum_{i=1}^M \pi_i{\cal N}(v|\mu_i, \Sigma_i)$, where $M$ is the number of mixtures for each base class. In order to satisfy the small memory footprint constraint, we use a GMM which assumes feature independence, \emph{i.e.,} the covariance matrix $\Sigma_i$ of each Gaussian mixture is diagonal. We denote this model as {\it Deep Feature GMM}. If we have $K$ classes, and the feature vectors dimensionality is $N$, the memory requirements for storing information about base classes is $O(MKN)$. The feature representation $v$, which we learn a generative model for, can be from the last fully connected layer or from deeper layers. In Section \ref{sec:deep_feature_gmm}, we evaluate the effectiveness of the use of the Deep Features GMM, showing that despite its compact representation, there is a minimal degradation in accuracy when training a classifier based only on data that is generated from the Deep Features GMM, compared to the accuracy obtained on the full training data.

\subsection{Low-Shot Training}
\label{training_procedure}
We apply standard data augmentation (random crop, horizontal flip, and color noise) to the input samples of the novel classes and create 100 additional samples variants from each of the novel class samples. These samples are passed through the feature extraction network $f_{rep}$ to obtain their corresponding feature representation. Note that new samples and their augmented variants are passed through $f_{rep}$ only once. 

As described in Section~\ref{single_layer}, we expand the classification subnetwork $f_{cls}$ and train the expanded network to classify novel classes in addition to the base classes. Figure~\ref{fig:Method}(a) illustrates the proposed method in the case where $f_{cls}$ is the last fully connected layer. 
As mentioned above, we only learn the $N$ dimensional vector $w_{K+1}$, which augments the $K\times N$ weight matrix of the FC layer. 

Each training batch is composed of base classes feature vectors drawn from the Deep Features GMM models learned from the base classes training data and the available samples of a novel class. The training batch is balanced to have an equal number of generations/samples per class. 

Since the forward and backward passes are carried out by only the last FC layers, each iteration can be done very rapidly. We use SGD with {\it gradient dropout} (see below) to learn $w_{K+1}$. More specifically, the weights update at step $(t+1)$ is done by:  
\begin{align*}
g_{K+1}^{(t+1)} &=  g_{K+1}^{(t)} + \mu \Delta w_{K+1}^{(t)} \\[7pt]
w_{K+1}^{(t+1)}&= w_{K+1}^{(t)} - \alpha M^{(t+1)} g_{K+1}^{(t+1)} 
\end{align*}
where $\mu$ is the momentum factor, $\alpha$  is the learning rate and $M^{(t+1)}$ is a binary random mask with probability $p$ of being $1$  ($M$ is randomly generated at each iteration throughout the low-shot training). That is, the gradient update is applied to a random subset of the learned weights. 
In Section \ref{sec:generic_cls} we demonstrate the contribution of the gradient dropout when only a few novel labeled samples are available.

\subsection{Expansion of Deeper Layers for Learning Representation}
\label{sec:extentions_and_representation}
The procedure described in the previous subsections expands the last classification layer but does not change the feature representation space. 
In some cases, especially in those which the novel classes are similar to the base classes, it is desirable to update and \emph{expand} the feature representation. 

To \emph{expand} the feature representation, we add new parameters to deeper layers of the network. This, of course, requires an appropriate expansion of all subsequent layers.  
To satisfy the hard distillation constraints, we enforce that the feature representation expansion does not affect the network output for the base classes. All weights in subsequent layers which connects the expanded representation to the base classes are set to zero and remain unchanged during learning. 
In Figure ~\ref{fig:Method}(b) we demonstrate an expansion of two last fully connected layers. The $FC2$  weight matrix is zero padded to adjust to the new added weights in $FC1$. Only the expansion to $FC2$ uses the newly added features in $FC1$.
The details of the representation learning expansion can be found in Supplementary Materials (Section S3).

\section{Experiments}
\label{sec:experiments}
In this section, we evaluate the proposed low-shot network expansion method on several classification tasks. We design a benchmark which measures the performance of several alternative low-shot methods in scenarios that resemble real-life problems, starting with easier tasks (Scenario 1) to harder tasks (Scenario 2 \& 3). In each experiment, we use a standard dataset that is partitioned into base classes and novel classes. We define three scenarios:
\begin{itemize}[]
\item[] 
	\textbf{Scenario 1, Generic novel classes:} unconstrained novel and base classes which can be from different domains.
\item[] 
	\textbf{Scenario 2, Domain specific with similar novel classes:} base and novel classes are drawn from the same domain and the novel classes share visual similarities among themselves.
\item[]
	\textbf{Scenario 3, Domain specific with similar base and novel classes:} base and novel classes are drawn from the same domain and each novel class shares visual similarities with one of the base classes.
\end{itemize}

In each scenario we define five base classes (learned using the full train set) and up to five novel classes, which should be learned from up to 15 samples only.  We compare the proposed method to several alternative methods for low-shot learning described in Section \ref{methods}. %
%
%


\subsection{Datasets for Low-Shot Network Expansion scenarios}
\label{dataset}
\paragraph{Dataset for Scenario 1}
\label{dataset:scenario_1_dataset}
For the task of generic classification of the novel classes, we use the ImageNet dataset \cite{ILSVRC15}, such that the selected classes were not part of the ILSVRC2012 1000 classes challenge. Each class has at least 1000 training images and 250 test images per class. We randomly selected 5 partitions of 5 base classes and 5 novel classes.

\paragraph{Dataset for Scenario 2 and Scenario 3}
\label{dataset:scenario23}
For these scenarios, we use the UT-Zappos50K \cite{ut-zap} shoes dataset for fine-grained classification. We choose 10 classes representing different types of shoes each having more than 1,000 training images and 250 test images. 

To define similarity between the chosen classes, we fine-tune the base network (VGG-19 \cite{VGG}) on the selected classes with the full dataset, and we use the confusion matrix as a measure of similarity between classes. Using the defined similarities, we randomly partition the 10 classes to 5 base and 2 novel classes, where for Scenario 2 we enforce similarity between novel classes, and for Scenario 3 we enforce similarity between novel and base classes. The confusion matrix is presented in Figure S2(b) in Supplementary Materials.
 
%

\subsection{Evaluated Methods} 
\label{methods}
In the proposed method we use the VGG-19 network \cite{VGG} trained on ImageNet ILSVRC2012 \cite{ILSVRC15} 1000 classes as a feature extraction subnetwork $f_{rep}$.
In all three scenarios for training the classification subnetwork $f_{cls}$ on the base classes, we fine-tune the last two fully-connected layers of VGG-19 on the 5 selected base classes, while freezing  the rest of the layers of $f_{rep}$. 

We denote the method proposed in Section \ref{sec:the-method} as Generative Low-Shot Network Expansion: \textit{Gen-LSNE}. We compare our proposed method to NCM \cite{mensink13pami}, and to the \textit{Prototype-kNN} method which is an extension of NCM and the soft distillation based method inspired by iCaRL method \cite{icarl}, adapted for the low-shot scenario. 

\subsubsection{NCM \& Prototype-kNN}
\label{method:knn}
We compare the proposed method to NCM classifier proposed by \mbox{
\cite{mensink13pami}}
. Additionally, we extend the NCM classifier by using multiple prototypes for each class, as in the {\it Prototype-kNN} classifier \cite{hastie01statisticallearning}. Both NCM and Prototype-kNN are implemented in a fixed feature space of the FC2 layer of the VGG-19 network. In our implementation of the Prototype-kNN, we fit a Deep Features GMM model with 20 mixtures for each of the base classes.  We extract feature representation of all of the available samples from the novel classes. The Deep Features GMM centroids of the base feature vectors and the novel feature vectors of the samples are considered as prototypes of each class. We set $k$ for Prototype-kNN classifier to be the smallest number of prototypes per class (the number of prototypes in the novel classes is lower than the number of mixtures in the base classes). 
The Prototype-kNN classification rule is the majority vote among $k$ nearest neighbors of the query sample. If the majority vote is indecisive, that is, there are two or more classes with the same number of prototypes among the $k$ nearest neighbors of the query image, we repeat classification with $k=1$.  

\subsubsection{Low-Shot with Soft Distillation}
We want to measure the benefit of the \emph{hard distillation} constraint in the low-shot learning scenario. Thus, we formulate a soft distillation based method inspired by iCaRL \cite{icarl} and methods described by \cite{venkatesan2017strategy} and \cite{Venkatesan:2016:NPL} as an alternative to the proposed method.

In the iCaRL method,  feature representation is updated by re-training the whole representation network. Since in low-shot scenario we have only a small number of novel class samples, updating the whole representation network is infeasible. Using the soft distillation method, we adapt to the low-shot scenario by updating only the last two fully connected layers $FC1,FC2$, but still use a combination of distillation and classification loss as in the iCaRL method.  

The iCaRL method stores a set of prototype images and uses the Nearest Mean Exemplar (NME) classifier at the final classification stage. In order to provide a fair comparison with the hard distillation method and uphold our memory restriction, we avoid storing prototypes in the image domain and use the proposed Deep-Features GMM as a generative model for the base-classes. 

To summarize, soft distillation applies a distillation loss and allows the $FC1,FC2$ layers to adjust to the new data, while the proposed hard-distillation freezes $FC1,FC2$ and trains only the new (expanded) parameters without using a distillation loss. We denote the soft distillation based methods as \textit{Soft-Dis} in the presented results.

\subsubsection{Gradient Dropout}
In Section~\ref{training_procedure} we proposed using gradient dropout regularization on SGD as a technique to improve convergence and overcome over-fitting in a low-shot scenario. We perform ablation experiments to assess the importance of the gradient dropout and train using both soft distillation (Soft-Dis) and proposed hard distillation (Gen-LSNE) with and without gradient dropout regularization. 

\subsection{\textbf{Results:} Expansion of the last fully connected layer}
\label{sec:generic_cls}
\label{sec:novelsim}

\begin{table}
\setlength{\tabcolsep}{10pt}
\begin{center}
\tiny
\begin{tabular}[b]{cccccc}
\multicolumn{6}{c}{Scenario 1 : Base + Novel Top-1 Test Error(\%)}\\
Method \slash NumSamples & 1&3&5&9&15\\
\hline
Prototype-KNN & 19.81&23.01&19.89&20.29&19.25\\
NCM & 21.3&9.84&8.89&7.92&7.71\\
\textit{Soft-Dis}+\textbf{GradDrop} & 21.46&12.04&9.45&7.48&\textbf{6.42}\\
\textit{Soft-Dis} & 21.31&12.41&9.82&7.48&6.5\\
\hdashline
Gen-LSNE+\textbf{GradDrop} & \textbf{15.21}&\textbf{9.82}&8.72&7.77&7.54\\
Gen-LSNE & 17.11&\textbf{9.82}&\textbf{8.46}&\textbf{7.15}&6.64
\end{tabular} \\

\begin{tabular}{c}
\tiny
\begin{tabular}[b]{cccccc}
\multicolumn{6}{c}{Scenario 2 : Base + Novel Top-1 Test Error(\%)}\\
Method \slash NumSamples & 1&3&5&9&15\\
\hline
Prototype-KNN & 34.12&40.08&33.08&32.95&30.65\\
NCM & 29.52&22.04&20.79&20.11&19.37\\
\textit{Soft-Dis}+\textbf{GradDrop} & 27.58&22.95&20.79&19.02&17.48\\
\textit{Soft-Dis} & 28.31&24.85&21.99&20.53&18.13\\
\hdashline
Gen-LSNE+\textbf{GradDrop} & \textbf{26.68}&21.15&20.58&19.36&18.5\\
Gen-LSNE & 27.06&\textbf{21.1}&\textbf{19.79}&\textbf{18.53}&\textbf{17.4}\\
\end{tabular} \\
\end{tabular}

\begin{tabular}{c}
\\
\begin{tabular}[b]{cccccc}
\multicolumn{6}{c}{Scenario 3 : Base + Novel Top-1 Test Error(\%)}\\
Method \slash NumSamples & 1&3&5&9&15\\
\hline
Prototype-KNN & 32.72&38.3&33.21&31.06&29.57\\
NCM & 29.48&21.98&22.78&21.53&20.79\\
\textit{Soft-Dis}+\textbf{GradDrop} & 30.17&24.0&22.64&20.86&18.4\\
\textit{Soft-Dis} & 30.22&24.45&23.36&20.98&18.78\\
\hdashline
Gen-LSNE+\textbf{GradDrop} & \textbf{24.23}&\textbf{20.44}&20.42&19.45&18.37\\
Gen-LSNE & 25.8&20.83&\textbf{20.35}&\textbf{19.39}&\textbf{17.77}\\

\end{tabular} \\
\end{tabular}

\end{center}
   \caption{\textbf{Expansion of last fully connected layer}: Top-1 Test Error on the proposed-method, Prototype-kNN, NCM and \textit{Soft-Dis}: the average Test Error on all 7 classes (base + novel).}
\label{table:LastFC}
\end{table}

\paragraph{Scenario 1: Generic novel classes}
In this experiment, the base classification network is trained on five base classes and then expanded to classify two novel classes chosen at random. For each of the five class partitions (Section~\ref{dataset}), we perform five trials by randomly drawing two novel classes from five novel classes available in the partition. The results are an average of 25 trials. The results of this experiment are presented in Table ~\ref{table:LastFC}(a). In Table~\ref{table:BaseNovelApart}(a) we present detailed results of the test error on the base and novels classes apart. Prototype-kNN and the \textit{Soft-Dis} methods perform better on the base classes. However, our method is significantly better on the novel classes and the overall test error is considerably improved, particularly when the number of samples is small. In addition, we see the significant gain in accuracy delivered by the gradient dropout when the number of novel samples is lower than 3 samples. Furthermore, gradient dropout also improves the results of the \textit{Soft-Dis} method.

NCM generally performs considerably better than Prototype-kNN in the Low-Shot scenario, despite the use of less information from the base classes. However, NCM is unable to effectively utilize more novel samples when they are available.  \method significantly outperforms NCM with a single novel sample, and overall outperforms all the tested method with nine and below samples per novel class.

\paragraph{Scenario 2 \& 3: Domain specific with similar novel-to-novel and novel-to-base classes} 
As described in Section ~\ref{dataset}.b, in each scenario we have 5 partitions with five base classes and two novel classes. The results are an average of 5 trials.
The result of the experiments are presented in Table \ref{table:LastFC}(b,c).
In Scenario-2~\&~Scenario-3 we see that the proposed method consistently outperforms the \textit{Soft-Dis}, NCM and Prototype-kNN methods. Training \method with gradient dropout improves results in cases with 1 \& 3 novel samples per class, especially in Scenario-3. 
In Table~\ref{table:BaseNovelApart}(b,c) we present detailed results of the test error on base and novels classes apart.

\subsection{\textbf{Results:} Expansion of Deeper Layers for Learning Representation}
\label{sec:represent}
In this section, we explore the effect of the expansion of deeper layers, as described in Section~\ref{sec:extentions_and_representation}. We partition the datasets as defined in \ref{dataset} to five base and five novel classes, and we test a 10 classes classification task. We expand the feature representation which is obtained after $FC1$ layer with 5 new features. The size of the feature representation after the FC1 layer of VGG-19 is of dimension 4k. Thus, $FC1$ is expanded with $4k\cdot 5$ new weights. The results are averaged over 5 trails (randomly selecting the base/novel classes). Table ~\ref{fig:exp5N} shows the results obtained, we denote \emph{+5Inner} as the experiments with the additional five shared representation features.


We see a marginal gain in Scenario~1. However, we observe a significant gain in Scenario~2 and~3 when the number of samples increases (especially Scenario~2). 

\begin{table}
\begin{center}
\begin{tabular}{c}
\multicolumn{1}{c}{\textbf{Scenario 1: Generic novel classes}}\\
\tiny
\begin{tabular}[b]{ccccccccc}
\multicolumn{9}{c}{Base + Novel Top-1 Test Error(\%)}\\
Method \slash \#Samples & 1&2&3&5&7&9&11&15\\
\hline
Prototype-KNN & 37.76&38.49&36.98&36.66&35.36&33.97&33.45&33.23\\
NCM & 36.08&22.79&17.18&15.54&14.96&13.7&13.37&13.17\\
\textit{Soft-Dis} & 37.39&26.34&20.46&15.6&14.04&12.18&11.4&10.62\\
\textit{Soft-Dis}+5Inner & 37.71&26.32&20.83&15.69&14.09&12.16&11.32&10.59\\
\hdashline
Gen-LSNE & \textbf{28.51}&\textbf{20.92}&17.15&14.55&13.35&11.98&11.27&10.9\\
Gen-LSNE+5Inner & 28.8&21.0&\textbf{17.14}&\textbf{14.46}&\textbf{13.15}&\textbf{11.7}&\textbf{11.13}&\textbf{10.44}\\

\end{tabular} \\
\multicolumn{1}{c}{(a)}\\
\multicolumn{1}{c}{\textbf{Scenario 2: Domain specific with similar novel classes}}\\
\tiny
\begin{tabular}[b]{ccccccccc}
\multicolumn{9}{c}{Base + Novel Top-1 Test Error(\%)}\\
Method \slash \#Samples & 1&2&3&5&7&9&11&15\\
\hline
Prototype-KNN & 41.28&40.37&39.63&39.96&39.11&37.31&38.06&36.42\\
NCM & 41.61&36.6&\textbf{32.74}&29.16&28.02&27.49&27.57&27.24\\
\textit{Soft-Dis} & 41.0&38.49&36.75&31.26&28.96&27.58&27.01&25.78\\
\textit{Soft-Dis}+5Inner & 41.3&38.43&37.18&31.53&28.98&27.39&27.06&25.57\\
\hdashline
Gen-LSNE & \textbf{38.52}&\textbf{35.95}&33.2&29.09&27.47&26.42&26.71&25.87\\
Gen-LSNE+5Inner & 39.11&36.45&33.95&\textbf{28.89}&\textbf{26.74}&\textbf{25.62}&\textbf{25.76}&\textbf{24.99}\\
\end{tabular}\\
\multicolumn{1}{c}{(b)}\\
\multicolumn{1}{c}{\textbf{Scenario 3: Domain specific with similar class in base}}\\
\tiny
\begin{tabular}[b]{ccccccccc}
\multicolumn{9}{c}{Base + Novel Top-1 Test Error(\%)}\\
Method \slash \#Samples & 1&2&3&5&7&9&11&15\\
\hline
Prototype-KNN & 48.69&46.81&48.7&47.96&45.02&43.72&44.17&44.35\\
NCM & 48.91&38.62&34.47&31.98&30.72&29.91&29.34&29.49\\
\textit{Soft-Dis} & 49.47&42.8&39.49&35.18&32.64&29.97&29.39&27.75\\
\textit{Soft-Dis}+5Inner & 49.52&42.64&39.45&35.35&32.42&30.22&29.36&27.81\\
\hdashline
Gen-LSNE & \textbf{41.0}&\textbf{34.29}&\textbf{32.5}&\textbf{29.93}&28.43&27.26&26.52&25.93\\
Gen-LSNE+5Inner & 42.07&34.37&33.45&30.07&\textbf{28.09}&\textbf{26.58}&\textbf{26.12}&\textbf{25.7}\\
\end{tabular}\\
\multicolumn{1}{c}{(c)}\\
\end{tabular}
\end{center}
   \caption{\textbf{Expansion of Deeper Layers for Learning Representation:} showing performance obtained with learning additional 5 shared inner features, \emph{+5Inner} marks the addition of the shared expanded features: (a) averaged results on Scenario 1  , (b) averaged results on Scenario 2, (c) averaged results on Scenario 3.}
\label{fig:exp5N}
\end{table}

\subsection{\textbf{Results:} Deep-features GMM Evaluation}
\label{sec:deep_feature_gmm}
In the Deep-features GMM evaluation experiment, we feed the full training data to the base network and collect the feature vectors before $FC1$, \emph{i.e.,} two FC layers before the classification output. We fit a GMM model to the feature vectors of each of the base classes with a varying number of mixtures. We train the two last FC layers of the base network from randomly initialized weights, where the training is based on generating feature vectors from the fitted GMM. We measure the top-1 accuracy on the test set of the networks trained with GMM models and the base network trained with full training data on the datasets defined in \ref{dataset}. 
%
%
The difference in top-1 accuracy between the network trained with full data and the networks trained with GMM models represent degradation caused by compressing the data with a simple generative model. The results of the experiment presented in the Table \ref{table:gmm} demonstrate that learning with samples from GMM models commonly causes only a negligible degradation relative to learning with a full training set. Together with the \emph{Hard-Distillation} constraint. Deep-Feature GMM is sufficient to imitate the presence of the inaccessible base class data. 

\begin{table}
\setlength{\tabcolsep}{8pt}
\begin{center}
\tiny
\begin{tabular}{l*{6}{c}r}
& \multicolumn{6}{c}{Top-1 Accuracy(\%)}\\
Dataset \slash  \# Mixtures& Full &1 & 10 & 20 & 40 & 60 \\
\hline

imagenet-group1-base&95.3&91.94&94.03&94.19&94.03&\textbf{94.57} \\
imagenet-group2-base&98.0&93.83&97.04&96.63&96.54&\textbf{97.37} \\
imagenet-group3-base&98.2&94.40&96.81&\textbf{97.45}&97.09&96.52 \\
imagenet-group4-base&98.8&95.60&98.16&98.01&98.30&\textbf{98.58} \\
imagenet-group5-base&99.0&97.26&98.26&98.01&98.01&\textbf{98.26} \\
ut-zap-scenario3-base&89.5&73.23&85.34&85.10&\textbf{85.50}&\textbf{85.50} \\
ut-zap-scenario2-novel&86.5&\textbf{81.81}&80.59&78.97&78.92&81.27 \\
ut-zap-scenario2-base&91.9&82.15&87.73&88.45&\textbf{91.16}&90.68 \\
\end{tabular}
\end{center}
   \caption{\textbf{Deep-Features GMM Evaluation:} Full stands for Fine tuning FC7,FC8 with the full training data.}
\label{table:gmm}
\end{table}

\section{Concluding Remarks}
\label{sec:summary}

We have introduced \method, a technique for low-shot network expansion. The method is based on hard-distillation, where pre-trained base parameters are kept intact, and only a small number of parameters are trained to accommodate the novel classes. We presented and evaluated the advantages of hard-distillation: (i) it gains significant increased accuracy (up to $20\%$) on the novel classes, (ii) it minimizes forgetting: less than $3\%$ drop in accuracy on the base classes, (iii) a small number of trained parameters avoids over-fitting, and (iv) the training for the expansion is fast. We have demonstrated that our method excels when only a few novel images are provided, rendering our method practical and efficient for a quick deployment of the network expansion. 

We have also presented Deep--Features GMM for effective base class memorization. This computationally and memory efficient method allows training the network from a generative compact feature-space representation of the base classes, without storing the entire training set. Finally, we have shown that the learned representation can be extended based on \emph{Low-Shot} novel observations to support better discrimination of novel classes.


In the future, we would like to continue exploring hard-distillation methods and extremely low-shot classifier expansion for robotic applications, 
aspiring towards human-level low-shot learning.

\section{Supplementary Materials}
\Urlmuskip=0mu plus 0mu
Supplementary materials can be found at: \footnotesize \url{https://github.com/adihayat/Gen-LSNE-Supplementary/blob/master/sup.pdf}


\begin{table}
\begin{center}
\smaller
\begin{tabular}{cc}

\tiny
\tabcolsep=0.14cm
\begin{tabular}[b]{cccccc}
\multicolumn{6}{c}{Base Top-1 Test Error(\%)}\\
Method \slash NumSamples & 1&3&5&9&15\\
\hline
Prototype-KNN & 6.27&3.01&\textbf{2.4}&\textbf{2.34}&\textbf{2.31}\\
NCM & 2.28&3.23&3.73&4.57&5.0\\
\textit{Soft-Dis}+\textbf{GradDrop} & \textbf{2.09}&\textbf{2.27}&2.43&2.7&2.98\\
\textit{Soft-Dis} & 2.21&2.36&2.41&2.65&2.85\\
\hdashline
Gen-LSNE+\textbf{GradDrop} & 2.64&3.75&4.3&4.84&5.43\\
Gen-LSNE & 2.34&2.96&3.5&3.99&4.33\\
\end{tabular} &

\tiny
\tabcolsep=0.10cm
\begin{tabular}[b]{ccccc}
\multicolumn{5}{c}{Novel Top-1 Test Error(\%)}\\
1&3&5&9&15\\
\hline
53.69&73.03&63.64&65.17&61.59\\
68.83&26.36&21.79&16.28&14.47\\
69.86&36.44&27.01&19.43&15.0\\
69.06&37.53&28.34&19.55&15.62\\
\hdashline
\textbf{46.62}&\textbf{25.0}&\textbf{19.75}&15.1&12.82\\
54.02&26.96&20.85&\textbf{15.03}&\textbf{12.43}\\

\end{tabular}\\

\multicolumn{2}{c}{(a)}\\

\end{tabular}
\\

\begin{tabular}{cc}

\tiny
\tabcolsep=0.10cm
\begin{tabular}[b]{cccccc}
\multicolumn{6}{c}{Base Top-1 Test Error(\%)}\\
Method \slash NumSamples & 1&3&5&9&15\\
\hline
Prototype-KNN & 19.2&23.05&18.08&18.35&15.61\\
NMC & 12.51&13.42&14.29&14.53&14.79\\
\textit{Soft-Dis}+\textbf{GradDrop} & \textbf{10.04}&\textbf{10.22}&\textbf{10.42}&\textbf{10.58}&\textbf{10.85}\\
\textit{Soft-Dis} & 11.41&11.82&11.6&12.22&11.73\\
\hdashline
Gen-LSNE+\textbf{GradDrop} & 10.57&11.75&12.57&12.88&13.24\\
Gen-LSNE & 10.44&10.98&11.87&11.97&12.33\\
\end{tabular} &

\tiny
\tabcolsep=0.10cm
\begin{tabular}[b]{ccccc}
\multicolumn{5}{c}{Novel Top-1 Test Error(\%)}\\
1&3&5&9&15\\
\hline
71.43&82.65&70.57&69.43&68.26\\
72.04&\textbf{43.6}&\textbf{37.02}&\textbf{34.06}&30.82\\
71.42&54.8&46.72&40.12&34.06\\
70.56&57.42&47.96&41.28&34.14\\
\hdashline
 \textbf{66.96}&44.66&40.58&35.58&31.64\\
68.62&46.4&39.58&34.94&\textbf{30.08}\\
\end{tabular}\\

\multicolumn{2}{c}{(b)}\\

\end{tabular}

\begin{tabular}{cc}

\tiny
\tabcolsep=0.10cm
\begin{tabular}[b]{cccccc}
\multicolumn{6}{c}{Base Top-1 Test Error(\%)}\\
Method \slash NumSamples & 1&3&5&9&15\\
\hline
Prototype-KNN & 17.83&16.22&13.48&11.49&10.63\\
NMC & 9.75&12.3&14.15&15.65&16.24\\
\textit{Soft-Dis}+\textbf{GradDrop} & 7.76&\textbf{7.89}&8.11&8.7&8.88\\
\textit{Soft-Dis} & 7.82&7.9&\textbf{8.1}&\textbf{8.53}&\textbf{8.75}\\
\hdashline
Gen-LSNE+\textbf{GradDrop} & 7.87&9.93&11.38&12.21&12.59\\
Gen-LSNE & \textbf{7.67}&8.7&10.29&10.88&11.57\\

\end{tabular} &

\tiny
\tabcolsep=0.10cm
\begin{tabular}[b]{ccccc}
\multicolumn{5}{c}{Novel Top-1 Test Error(\%)}\\
1&3&5&9&15\\
\hline
69.93&93.5&82.52&79.98&76.91\\
78.8&\textbf{46.18}&44.34&\textbf{36.23}&\textbf{32.16}\\
86.2&64.29&58.97&51.25&42.21\\
86.23&65.81&61.52&52.11&43.85\\
\hdashline
\textbf{65.12}&46.72&\textbf{43.04}&37.57&32.82\\
71.13&51.15&45.49&40.64&33.26\\

\end{tabular}\\

\multicolumn{2}{c}{(c)}\\

\end{tabular}
\end{center}
   \caption{\textbf{Base \& Novel Accuracy Apart}: Top-1 average test error rate on the proposed-method, Prototype-kNN, NCM and \textit{Soft-Dis} \textbf{base} classes (5 from 7) and \textbf{novel} classes (2 from 7) apart: (a) Scenario 1, Generic novel classes (b) Scenario 2, Domain specific with similar novel classes (c) Scenario 3, Domain specific with similar base and novel classes.}
\label{table:BaseNovelApart}
\end{table}

\bibliographystyle{IEEEtran}
\bibliography{lowshotbib}




\end{document}